\documentclass[letterpaper, 10 pt, conference]{ieeeconf}

\IEEEoverridecommandlockouts
\usepackage{amsmath}
\usepackage{amssymb}
\usepackage{capt-of}
\usepackage{graphicx}
\usepackage{url}
\makeatletter
\let\NAT@parse\undefined
\makeatother
\usepackage[colorlinks=true]{hyperref}
\usepackage[all]{hypcap}
\usepackage{cleveref}
\crefname{figure}{Fig.}{Figs.}
\Crefname{figure}{Figure}{Figures}

\renewcommand{\footnoterule}{%
  \kern -3pt
  \hrule width 0.4\columnwidth
  \kern 2.6pt
}

\title{Training in Deployment: Human-in-the-loop Online Post-Training for Robot Manipulation}

\author{\parbox{\textwidth}{\centering
  \normalsize
  Chun Gu$^{1,2,3*}$, Xiaofei Wei$^{1,2,3*}$, Yiming Chen$^{2,3}$,
  Bozhou Zhang$^{1,3}$, Juntong Zhang$^{2,3}$, Xiatian Zhu$^{4}$, Li Zhang$^{1,2,3\dagger}$\\[0.4em]
  \small
  $^{1}$School of Data Science, Fudan University\quad
  $^{2}$Shanghai Innovation Institute\quad $^{3}$Logos Intelligence Lab\\
  $^{4}$University of Surrey\\[0.9em]
  \normalfont\href{https://logosintelligencelab.github.io/TIDE/}{\nolinkurl{logosintelligencelab.github.io/TIDE}}
}\thanks{$^{*}$Equal contribution.}%
\thanks{$^{\dagger}$Corresponding author (\href{mailto:lizhangfd@fudan.edu.cn}{\nolinkurl{lizhangfd@fudan.edu.cn}}).}}

\hypersetup{
  pdftitle={Training in Deployment: Human-in-the-loop Online Post-Training for Robot Manipulation},
  pdfauthor={Chun Gu, Xiaofei Wei, Yiming Chen, Bozhou Zhang, Juntong Zhang, Xiatian Zhu, Li Zhang}
}

\IEEEaftertitletext{\begin{minipage}{\textwidth}
  \centering
  \makeatletter\def\@captype{figure}\makeatother
  \capstart
  \includegraphics[width=0.95\textwidth]{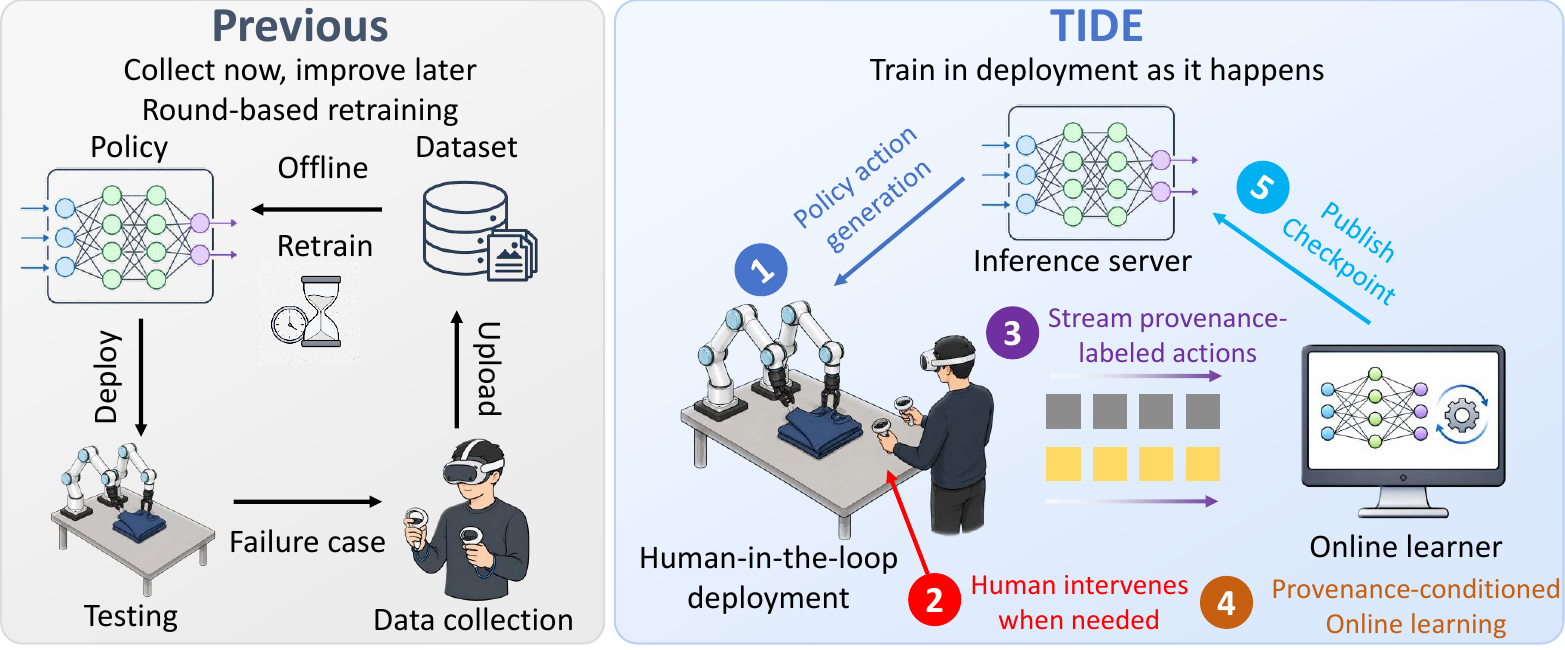}
  \captionof{figure}{\textbf{Training in Deployment.}
  Left: Round-based methods collect data for later offline retraining.
  Right: TIDE learns online from human interventions and policy executions,
  then redeploys updated checkpoints, making policy improvements available
  within the same deployment session.}
  \label{fig:teaser}
  \vspace{5mm}
\end{minipage}
}

\begin{document}

\maketitle
\thispagestyle{empty}
\pagestyle{empty}

\begin{abstract}
Large pretrained robot foundation models provide strong priors for manipulation, yet adapting them to target tasks with offline teleoperated demonstrations often fails to cover the contact-rich, recovery, and precision-critical states encountered during deployment. Existing methods mitigate this issue by augmenting offline demonstrations with human-in-the-loop data or online reinforcement learning. Despite this, offline post-training adapts slowly because it cannot immediately absorb deployment experience, while online RL typically relies on value functions or auxiliary actor-critic models, increasing both training complexity and instability. We present \textbf{TIDE} (Training In DEployment), a human-in-the-loop online post-training system. TIDE introduces a provenance-conditioned supervised learning framework that updates the policy online from human interventions and policy-generated rollouts collected during deployment, without a learned critic, reward model, or action scorer. Across four real-world bimanual manipulation tasks, TIDE improves milestone-normalized task progress and binary full-task completion over the warm-start policy and matched additional offline demonstrations, and performs comparably to the fixed-policy collection baseline while making updated checkpoints available during deployment. Each task-specific online post-training session, from the start of deployment rollouts until the final checkpoint is available, completes in less than one hour while learning and checkpoint handling run asynchronously.

\end{abstract}


\section{Introduction}
\label{sec:introduction}
Modern robot manipulation increasingly relies on large pretrained robot foundation models~\cite{black2024pi_0,kim2024openvla,team2024octo,zitkovich2023rt,ye2026world,shen2026videovla}. A common post-training paradigm adapts these pretrained priors to a target task using offline human teleoperation data~\cite{black2024pi_0,zhao2024aloha}, producing an initially deployable policy. However, real-robot failures often arise in off-demonstration states induced by the policy itself, including contact-rich interactions, recovery situations, and precision-critical configurations. Covering such rare but deployment-critical states by simply collecting additional offline teleoperation episodes is inherently data-inefficient~\cite{ross2011reduction,hu2025rac,zhao2024aloha}.

Human-in-the-loop learning offers a natural mechanism for covering these policy-induced states, as human operators can intervene during failures or challenging recovery behaviors. However, existing approaches often treat deployment primarily as a provider of expert action labels for supervised learning: human-intervention segments are used as demonstrations, while policy-generated segments are discarded or treated as unlabeled data~\cite{ross2011reduction,kelly2019hg,mandlekar2020human,wu2025robocopilot}. This limits data efficiency, especially when most deployment experience is generated by the policy itself. Moreover, offline-only post-training methods~\cite{kelly2019hg,mandlekar2020human,liu2025robot,wu2025robocopilot,intelligence2025pi} typically follow a collect-train-deploy cycle, preventing them from immediately incorporating deployment experience and thereby slowing adaptation under distribution shift.
Online reinforcement learning leverages both human interventions and policy-generated rollouts, bringing policy updates closer to deployment~\cite{xu2026rl,luo2025precise}. However, these methods often rely on value functions or critics trained from sparse episode-level success-or-failure feedback. Such weak supervision makes it difficult to stably evaluate high-dimensional action chunks, increasing both training complexity and instability. Other systems reduce this burden by updating only small policy modules or by shifting annotation away from real-robot rollouts~\cite{xu2026rl,fang2026robopocket,lu2025human}. While effective in constrained settings, these design choices limit the potential of online post-training to improve large robot foundation models across diverse manipulation tasks.
Existing methods therefore lack a mechanism for promptly incorporating deployment-time intervention and execution signals without deferring learning to a subsequent offline training round or relying on difficult-to-estimate value functions.

Our central insight is that, once a pretrained policy is warm-started on a target task, deployment need not be a data-collection stage before the next offline retraining round. It can instead serve as a continual learning process that produces structured training signals. Human interventions and policy executions reveal what the current policy can accomplish and which state regions still require repair. Deployment data should therefore be organized as provenance-structured online experience rather than flattened into additional expert demonstrations.

Building on this insight, we introduce \textbf{TIDE} (Training In DEployment), a human-in-the-loop online post-training system for pretrained manipulation policies. As shown in \cref{fig:teaser}, TIDE places the robot, the human operator, and the learner in a single deployment loop: the robot executes action chunks generated by the current policy, the operator intervenes only when necessary, and the system records the executed action at each step together with its provenance, i.e., whether it was produced by the human or the policy. Provenance records execution source only; it is not a reward, correctness label, or quality score. The learner continuously updates all trainable policy modules using both offline task data and the growing online dataset, and returns updated checkpoints to the robot at episode boundaries. Rather than treating mixed online episodes as homogeneous expert demonstrations, TIDE restructures them into pure-provenance chunks. These labels enable provenance-conditioned supervised training without a separate reward model, critic, or action scorer.

We validate TIDE on four real-robot manipulation tasks spanning precision contact, deformable-object manipulation, constrained placement, and recovery from policy-induced states. Across these tasks, TIDE improves no-takeover task performance over the warm-start policy and matched additional offline demonstrations. It performs comparably to the Fixed Policy baseline while making updates available during deployment. The observed intervention ratio also decreases over deployment, indicating lower human-supervision burden without serving as an autonomous-performance measure.

This paper makes three main contributions: (1) \textbf{TIDE}, a human-in-the-loop online post-training system that keeps real-robot execution, human takeover, learning, and policy redeployment in one session; (2) provenance-conditioned joint optimization of all trainable policy modules, turning mixed human-policy deployment data into structured source supervision; and (3) real-robot evidence across four manipulation tasks showing improved no-takeover task performance and a declining observed intervention ratio.

\section{Related work}
\label{sec:related-work}

\begin{figure*}[t]
  \centering
  \includegraphics[width=\textwidth]{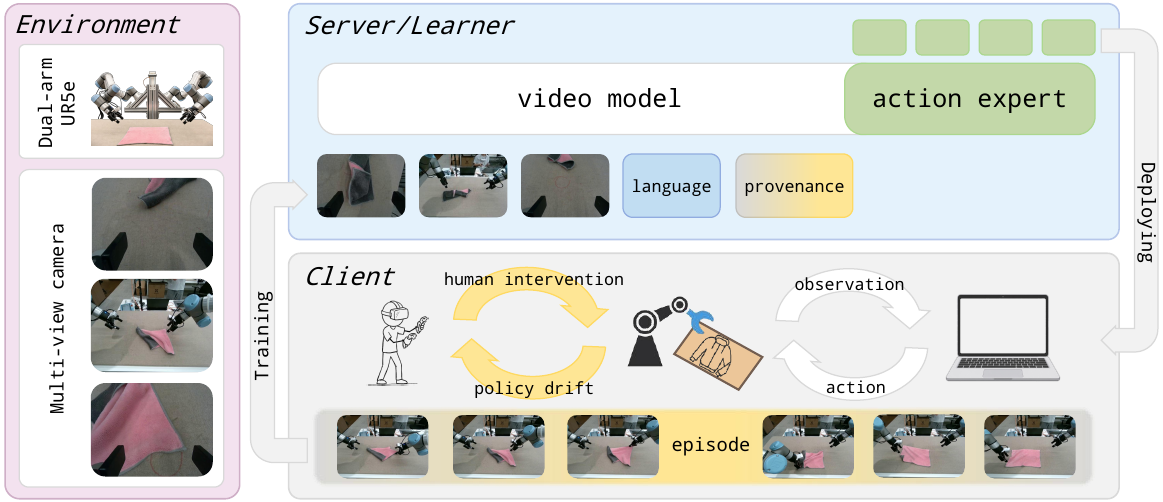}
  \caption{\textbf{The pipeline of TIDE.} The robot executes policy actions with
  human takeover when needed. The system records executed actions and labels
  their human or policy source. The learner combines this online data with
  offline demonstrations to update the video model and action expert using
  provenance conditioning. Updated policies return to the robot at episode
  boundaries, while data collection and learning proceed asynchronously
  within the same deployment session.}
  \label{fig:pipeline}
\end{figure*}

\subsection{Robot foundation models for manipulation}
Modern manipulation policies are increasingly built on large
pretrained robot foundation models that map observations and language
instructions to actions~\cite{brohan2022rt,zitkovich2023rt,
kim2024openvla,team2024octo,black2024pi_0,team2025gemini,
bjorck2025gr00t}.
RT-1 and RT-2~\cite{brohan2022rt,zitkovich2023rt},
OpenVLA~\cite{kim2024openvla}, and Octo~\cite{team2024octo}
established this recipe for broad instruction-conditioned
manipulation, while recent generalist and humanoid systems extend it
across embodiments and physical settings~\cite{team2025gemini,
bjorck2025gr00t}.
Progress has also come from more capable action interfaces, including
flow-matching action experts over continuous action
chunks~\cite{black2024pi_0,intelligence2025pi,zhao2023,chi2025diffusion},
and efficient action tokenization~\cite{liu2025rdt,
pertsch2025fast}.
In parallel, models that couple action generation with future visual
state prediction use video-generation priors to improve robustness to
unfamiliar skills and embodiments~\cite{ye2026world,shen2026videovla,
guo2026unified,pai2025mimic,du2023learning,wu2024unleashing,
hu2024video,tian2025predictive}.
These advances rely on increasingly large cross-embodiment datasets
for robot learning~\cite{khazatsky2024droid,walke2023bridgedata,
o2024open,bu2025agibot}.
Yet deployed policies remain bounded by training-distribution
coverage: small errors compound into out-of-distribution states, and
precision-critical subtasks can resist improvement even after many
additional teleoperation episodes~\cite{hu2025rac,zhao2024aloha}.
This limitation motivates post-training procedures that refine
pretrained policies through interaction with the deployment
environment.

\subsection{Human-in-the-loop post-training}

Human-in-the-loop learning extends imitation learning by allowing a
human to intervene on states visited by the current policy, directly
addressing covariate shift and compounding
errors~\cite{ross2011reduction,laskey2017dart,kelly2019hg}.
Subsequent systems improve when interventions are triggered, how mixed
demonstration and intervention data are weighted, what corrective data
are collected, or how disruptive takeover
is~\cite{mandlekar2020human,liu2025robot,liu2024multi,hu2025rac,
wang2025genie,hoque2024intervengen,xu2026compliant,zhang2016query,hoque2021thriftydagger,hoque2021lazydagger,menda2019ensembledagger}.
Related methods extract different learning signals from interventions:
RLIF treats intervention signals as rewards for off-policy reinforcement
learning, MILE models the intervention process and uses both intervention
and non-intervention feedback, and Sirius reweights deployment data with
approximated human trust for weighted behavioral
cloning~\cite{luo2024rlif,korkmaz2025mile,liu2025robot}.
Many interactive imitation-learning systems still update models in batches.
RoboCopilot~\cite{wu2025robocopilot} moves toward continual
fine-tuning within an interactive collection loop, but its workflow
remains round-based and the deployed policy still comes from batched
retraining.
RECAP~\cite{intelligence2025pi} scales full-model HIL improvement with
advantage conditioning over demonstrations, rollouts, and corrections,
but it still operates through phase-based offline updates and leaves a
fully concurrent online framework as future work.

Recent systems move policy improvement closer to deployment, but each
keeps only part of the desired closed loop. RL
Token~\cite{xu2026rl} performs online learning on top of a frozen
pretrained policy by training a lightweight actor-critic rather than
updating the full backbone. Value-based routes must estimate action
quality from sparse outcome feedback~\cite{xu2026rl,wang2026learning,
luo2024serl,chebotar2023q},
which is difficult for high-dimensional action chunks in real robots.
Hi-ORS~\cite{lu2025human}
avoids a learned critic through outcome-based rejection sampling and
supervised updates, but still relies on trajectory-level outcome
supervision; ConRFT~\cite{chen2025conrft} instead retains a Q-function
and gates its online consistency-policy fine-tuning with human
interventions. RoboPocket~\cite{fang2026robopocket} accelerates policy
iteration by visualizing predicted policy trajectories in AR and
collecting human corrections without executing those predictions on
the physical robot.

TIDE targets a complementary setting in which real-robot deployment
remains in the data loop, the video backbone, cross-modal adapter, and
action expert are jointly updated, and supervision is organized by
action-source provenance rather than learned values or trajectory-level
reward labels.

\section{Method}
\label{sec:method}

TIDE builds on a pretrained world-action policy consisting of a video-generation backbone $\Phi$ and a flow-matching action expert $v_\theta^A$. Given an observation $o_t$ and a language instruction $q$, the backbone produces a compact world prior $z_t = \Phi(o_t, q)$. The action expert cross-attends to $z_t$ and the robot state $s_t$ to generate a 50-step action chunk $A_t = (a_t, \dots, a_{t+49})$, while the backbone predicts future video frames over the same horizon. Before deployment, we warm-start all trainable policy modules on offline human teleoperation data for the target task.

This warm-started policy provides an executable initialization, but deployment can still expose states that are not covered by offline demonstrations. TIDE turns these deployment-time states into training signals within the same session. We first describe the training-in-deployment runtime, which integrates robot execution, provenance recording, online learning, and checkpoint handoff in a single loop (Section~\ref{subsec:runtime}). We then explain how execution records are converted into provenance-labeled chunks for provenance-conditioned training (Section~\ref{subsec:learning}). An overview of the pipeline is shown in \cref{fig:pipeline}.

\begin{figure*}[t]
  \centering
  \includegraphics[width=\textwidth]{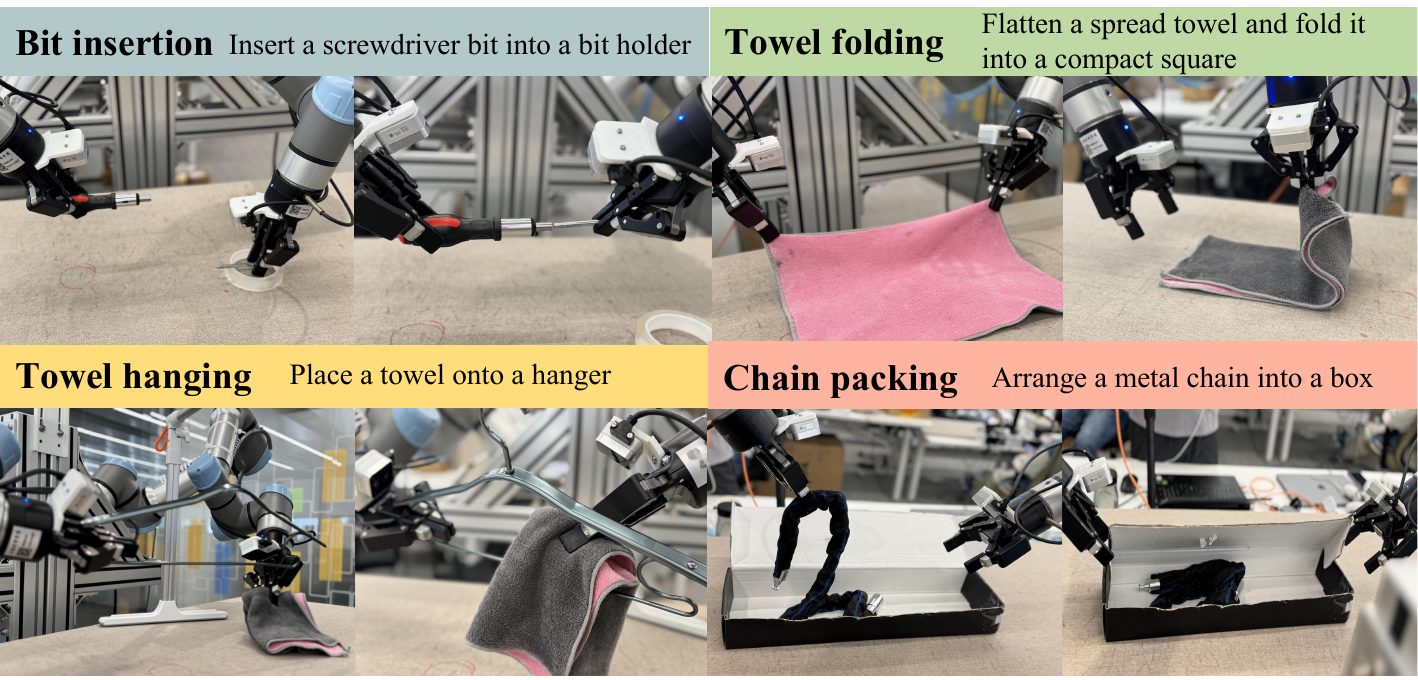}
  \caption{\textbf{Real-world evaluation tasks.} Initial and successful final
  states for four bimanual manipulation tasks.}
  \label{fig:tasks}
\end{figure*}

\subsection{Training-in-deployment system}
\label{subsec:runtime}

The training-in-deployment system couples a real-robot client, an inference-and-recording server, and an online learner within a single deployment session. The client executes actions from either the policy or the human operator and streams executed-action records to the server. The server handles policy inference, stores the incoming records, and makes them available to the learner. The learner updates all trainable policy modules online and returns new policy checkpoints at episode boundaries.

\paragraph{Robot client}
The robot client is responsible for low-latency real-robot control. At each control step, it collects observations, requests action chunks from the server, and executes policy actions unless the operator intervenes. Inspired by RTC~\cite{black2025real,black2025training}, the client adopts asynchronous real-time chunking: committed actions continue to execute while future actions are refreshed in the background. This design reduces inference-induced stalls and smooths transitions between consecutive action chunks. During takeover, VR delta control allows the operator to command incremental motions around the current robot state and return control to the policy without resetting the episode. Each handoff clears queued policy chunks to prevent stale actions from being executed. The client streams executed actions and intervention flags asynchronously to the server, keeping data recording off the critical control path.

\paragraph{Inference and recording server}
The inference-and-recording server mediates between real-robot control and online training. It centralizes GPU-based policy inference and writes online episodes to a shared dataset, while leaving low-latency control to the robot client. For each client query, the server runs the active checkpoint on the current observation and returns an action chunk. In parallel, it appends execution records and manages episode commit or discard. At episode end, the human operator manually confirms task completion. Before upload, the server commits only confirmed-successful episodes; failed, timed-out, aborted or stopped, reset-before-completion, re-recorded, stale, and unrecovered episodes are discarded. To ensure rollout consistency, the server adopts learner-produced checkpoints only at episode boundaries, so that each rollout is associated with a single policy generation.

\paragraph{Online learner}
The online learner continuously incorporates deployment experience into the pretrained policy during the same deployment session. As committed records arrive, it reads from the growing online dataset and jointly updates the video diffusion backbone, cross-modal adapter/Q-former, action expert, and source embeddings with offline task data; the pretrained text encoder and video VAE remain frozen. The learner allocates 0.2 optimizer steps per newly committed control frame and publishes a checkpoint every five committed episodes. The server prefetches new weights and adopts them only at the next episode boundary. This mechanism returns model improvements to the robot during deployment while ensuring that each rollout is generated by a single policy generation.

\subsection{Provenance-conditioned learning}
\label{subsec:learning}

\paragraph{Provenance-labeled chunks}
The execution loop described above produces a continuous stream of mixed deployment episodes containing both human interventions and policy executions. To convert these episodes into supervision, TIDE assigns provenance labels at the action-chunk level rather than treating an entire episode as a single demonstration. Human-executed chunks provide supervision for failure recovery and difficult states, while policy-executed chunks provide additional learning signal from behaviors that the current policy can already perform. Chunk-level provenance labels allow TIDE to use both sources of deployment data without flattening mixed episodes into homogeneous expert demonstrations.

The provenance labels are derived directly from signals recorded during execution. At each control step, TIDE stores the action sent to the robot and whether that action was produced by the human operator or by the policy. Provenance encodes action-source identity only, not action quality. Episode success is used only to decide whether an episode enters the online training dataset; it is not a model condition, reward, provenance label, or training target. Given an action chunk $A$, TIDE assigns
\begin{equation}
\label{eq:provenance-label}
\ell(A)=
\begin{cases}
0, & A \text{ is fully human-executed},\\
1, & A \text{ is fully policy-executed},\\
\varnothing, & A \text{ mixes human and policy steps}.
\end{cases}
\end{equation}
A 50-step target-action horizon receives a human or policy label only when all 50 actions share that source. Horizons labeled $\varnothing$ contain both sources and are excluded from provenance-conditioned training, with no additional exclusion window around a handoff.

\paragraph{Provenance-conditioned online training}
Given the provenance-labeled chunks defined above, TIDE trains all trainable policy modules with a provenance-conditioned flow-matching objective. We map each provenance label $\ell$ to a learnable embedding $c_\ell$ and condition both the video-generation backbone and the action expert on this embedding. The backbone takes $(o_t, q, c_\ell)$ as input to produce a provenance-conditioned world prior $z_\ell = \Phi(o_t, q; c_\ell)$ and predict the future-video velocity $v_\phi^I$. The action expert then conditions on $(z_\ell, s_t, c_\ell)$ to predict the action velocity $v_\theta^A$. In both networks, $c_\ell$ is added to the flow-time embedding. All reported online autonomous rollouts and final no-takeover evaluations fix the source condition to the human mode, $\ell=0$.

We use a linear-Gaussian probability path. Given a flow time $\tau \in [0,1]$ and independent noises $\varepsilon_I, \varepsilon_A \sim \mathcal{N}(0,\mathbf{I})$, we construct noisy interpolants
$I_t^\tau = \tau I_t + (1-\tau)\varepsilon_I$ and
$A_t^\tau = \tau A_t + (1-\tau)\varepsilon_A$, with target velocities
$I_t - \varepsilon_I$ and $A_t - \varepsilon_A$, respectively. The training objective is
\begin{equation}
\label{eq:provenance-conditioned-loss}
\begin{split}
\mathcal{L}_{\mathrm{WAM}}=\mathbb{E}\Big[&
\bigl\| v_\phi^I(I_t^\tau, \tau \mid o_t, q, c_\ell)
      - (I_t - \varepsilon_I) \bigr\|_2^2 \\
&+\bigl\| v_\theta^A(A_t^\tau, \tau \mid z_\ell, s_t, c_\ell)
      - (A_t - \varepsilon_A) \bigr\|_2^2
\Big].
\end{split}
\end{equation}

The training data combine offline teleoperation data, treated as
human-provenance chunks with $\ell=0$, and the pure-provenance online
chunks described above. The learner maintains separate offline and online
buffers. Online samples receive a 5:1 weight relative to offline samples,
corresponding to selection probabilities $5/6$ and $1/6$ between the two
buffers. Preliminary tuning selected $\alpha=5$, which is fixed across all
reported tasks and variants. Within the online buffer,
human-executed and policy-executed chunks are balanced so that sparse
intervention data are not overwhelmed by autonomous execution.

We also apply two task-preserving data augmentations while jointly optimizing all trainable policy modules.
Space mirroring reflects observations, robot states, and
actions across the workspace symmetry plane when the task remains valid
under this transformation, increasing coverage of left and right spatial
configurations. Time scaling resamples action chunks and corresponding
future observations at different execution rates, exposing the model to
moderate speed variation. Each augmented sample inherits the provenance
label of its original chunk.

\section{Experiments}
\label{sec:experiments}

\subsection{Experimental setup}
\label{subsec:exp-setup}

\begin{figure}[t]
  \centering
  \includegraphics[width=\linewidth]{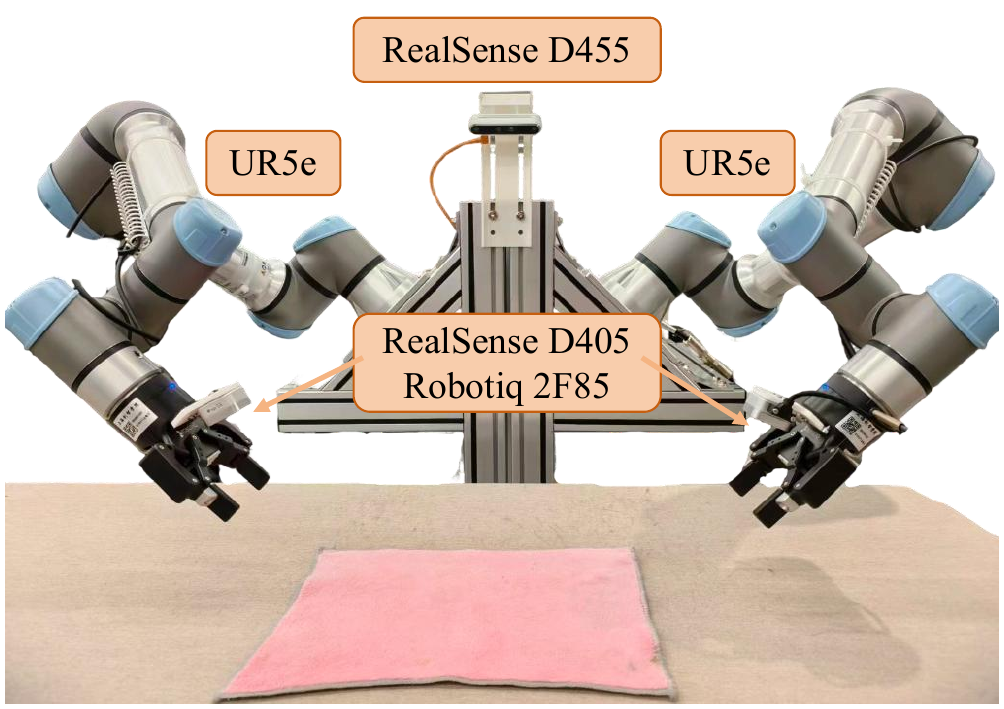}
  \caption{\textbf{Real-robot setup.} The bimanual UR5e platform uses
  Robotiq 2F85 grippers,
  one third-person RealSense D455 camera, and two wrist-mounted
  RealSense D405 cameras.}
  \label{fig:robot-setup}
\end{figure}

\paragraph{Robot platform}
We conduct all real-world experiments on the bimanual UR5e platform
shown in \cref{fig:robot-setup}. The perception setup includes one
third-person D455 camera and two wrist-mounted D405 cameras, providing
both scene-level context and local contact observations near the
grippers.

\paragraph{Tasks}
We evaluate TIDE on four real-world bimanual manipulation tasks shown in
\cref{fig:tasks}. \textbf{Bit Insertion} requires inserting a
screwdriver bit into a bit holder and tests precise pose control under
local contact. \textbf{Towel Folding} requires flattening a spread towel
and folding it into a compact square, stressing long-horizon
deformable-object manipulation. \textbf{Towel Hanging} requires placing
a towel onto a hanger, combining deformable manipulation with a
constrained target. \textbf{Chain Packing} requires arranging a metal
chain into a box, creating recovery states from local deformation and
possible tangling.

\paragraph{Protocol and metrics}
Every method--task pair is evaluated using 20 autonomous rollouts from
the same predefined initial-state set, with no human takeover. TIDE uses
the human source condition $\ell=0$ during these evaluations. Let
$c_{r,t}$ denote the number of milestones completed in rollout $r$ for
task $t$, and let $C_t$ denote the total number of milestones for that
task. We compute the normalized rollout score and aggregate task progress as
\[
    a_{r,t}=\frac{c_{r,t}}{C_t}, \qquad
    \bar{a}_t=\frac{1}{N}\sum_{r=1}^{N}a_{r,t}, \quad N=20.
\]
If a rollout ends because of a timeout, safety stop, or task failure,
only milestones already stably completed are counted. We report binary
full-task completion separately as $k/20$. A rollout is complete only if
$c_{r,t}=C_t$. During HIL collection, only episodes manually
confirmed as successful are committed to the online training dataset;
success affects dataset inclusion only. We define human intervention
ratio as the fraction of executed control steps under human takeover and
use it only as a within-task measure of supervision burden.

The task milestones are pick/insert for Bit Insertion;
flatten/first fold/second fold for Towel Folding; pick hanger/place
towel/hang hanger for Towel Hanging; and place chain/close lid/tuck
flaps for Chain Packing.

\paragraph{Baselines}
Every method starts from the same task-specific warm-start checkpoint.
The offline-demo baseline uses newly collected teleoperation
demonstrations matched to the amount of online data, then trains for the
same number of optimizer steps as TIDE. The fixed-policy collection baseline (Fixed Policy)
holds the warm-start policy fixed during HIL collection matched to TIDE
by duration, then trains after collection using the same objective,
buffer configuration, filtering rules, and total optimizer-update
budget. TIDE instead trains and redeploys checkpoints during collection.

\paragraph{Implementation details}
The reported world-action policy has 3B parameters and is trained on
eight H100 GPUs. Each online optimizer step takes approximately 0.4\,s
on this setup. Online training jointly updates the video diffusion
backbone, cross-modal adapter/Q-former, action expert, and source
embeddings, while the text encoder and video VAE remain frozen. The
policy predicts a 50-step, 14-dimensional bimanual joint-relative action
chunk. Each predicted offset is added to the current joint state before
the controller receives the resulting absolute joint target. Online
samples have a 5:1 weight relative to offline samples ($\alpha=5$),
equivalent to an online-buffer selection probability of $5/6$; this
preliminarily tuned setting is fixed across all reported tasks and
variants. The learner takes 0.2 optimizer steps per newly committed
control frame and publishes a checkpoint every five committed episodes;
the server prefetches and adopts it at the next episode boundary. For
each task, the continuous wall-clock interval from the start of online
HIL rollouts until the final checkpoint is deployable is less than one
hour. This interval includes asynchronous data processing, training,
checkpoint handling, robot execution, human takeover, and episode resets,
but excludes initial setup and final autonomous evaluation. Online
training runs concurrently with data collection and episode resets, and
its wall-clock cost is included in this reported interval. For Bit
Insertion, collection comprised 41 committed episodes and 38,839 control
frames (21.6 minutes at 30 Hz): 1,432 human-controlled frames (3.69\%)
and 37,407 policy-controlled frames.

\subsection{Real-world experiments}
\label{subsec:real-robot-evaluation}

\begin{figure}[t]
  \centering
  \includegraphics[width=\linewidth]{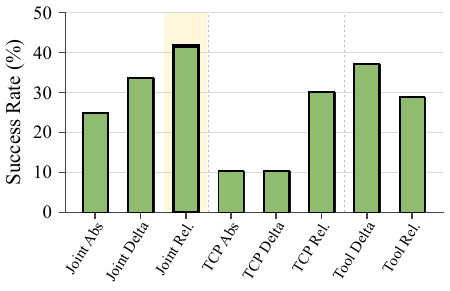}
  \caption{\textbf{Action representation study.} Offline-only comparison of
  eight action spaces using the same data, initialization, and training budget.
  Joint-relative actions perform best and are used in subsequent experiments.}
  \label{fig:action-space}
\end{figure}

\paragraph{Q1: Which action representation should be used}
Action representation affects how the policy expresses contact-rich
motions, so we first evaluate this choice before online post-training.
We train offline-only policies with the same data, initialization, and
budget, comparing eight spaces: absolute, delta, and relative actions in
joint and TCP space, plus delta and relative actions in tool space. As
shown in \cref{fig:action-space}, joint-relative actions perform best
in this controlled study and are used in all subsequent experiments.

\begin{figure}[t]
  \centering
  \includegraphics[width=\linewidth]{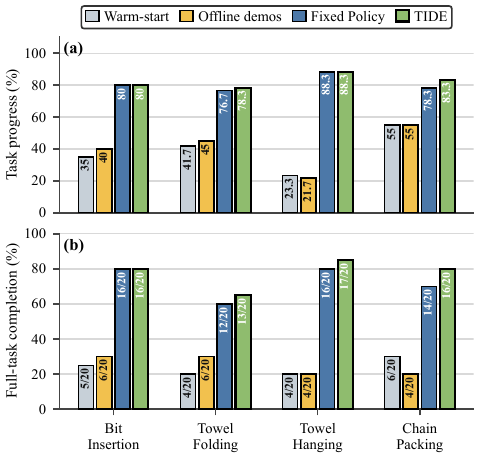}
  \caption{\textbf{Autonomous evaluation over 20 no-takeover rollouts per
  method--task pair.} \textbf{(a)} Mean milestone-normalized task progress.
  \textbf{(b)} Binary full-task completion. Labels report completed rollouts
  out of 20.}
  \label{fig:success-bars}
\end{figure}

\begin{figure*}[ht]
  \centering
  \includegraphics[width=\linewidth]{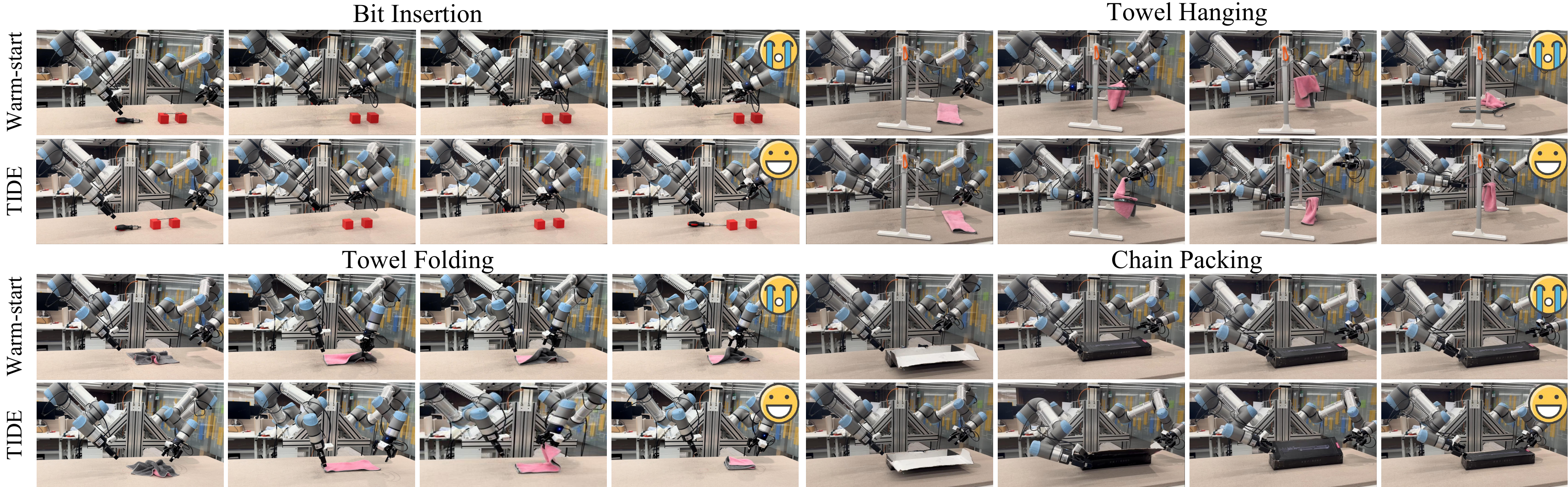}
  \caption{\textbf{Qualitative autonomous rollouts before and after online post-training.}
  For each task, the top and bottom rows show no-takeover rollouts from the
  warm-start policy and final TIDE checkpoint, respectively, with time
  progressing from left to right. The warm-start policy fails to complete the
  tasks in these examples, whereas TIDE completes the full manipulation
  sequences, illustrating improved autonomous execution after online
  post-training.}
  \label{fig:qualitative-autonomous-rollouts}
\end{figure*}

\paragraph{Q2: Does TIDE improve autonomous task performance}
Because human takeover can complete training episodes that would
otherwise fail, we measure task performance only with the no-takeover
evaluation described above. \Cref{fig:success-bars} reports
milestone-normalized task progress and binary full-task completion. The
completion labels give the number of completed rollouts out of 20. The two
metrics capture different outcomes. Progress credits intermediate milestones.
Completion requires the entire task. Both
Fixed Policy and TIDE
numerically outperform Warm and Offline on milestone-normalized progress
for all four tasks. TIDE matches Fixed Policy on Bit Insertion and Towel Hanging
progress, and is higher on Towel Folding and Chain Packing. Binary
completion is tied on Bit Insertion and is higher for TIDE by one rollout
on both towel tasks and by two rollouts on Chain Packing. Thus, Fixed Policy and
TIDE have comparable final performance under this evaluation. TIDE's
distinction is that updated checkpoints become available during the
deployment session rather than only after a separate collection,
training, and redeployment cycle.
\Cref{fig:qualitative-autonomous-rollouts} provides illustrative
before-and-after rollouts; the quantitative comparison is based on the
20-rollout evaluations above.

\begin{figure}[t]
  \centering
  \includegraphics[width=\linewidth]{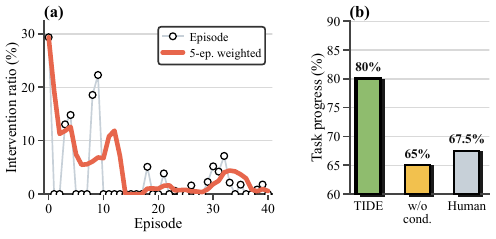}
  \caption{\textbf{Deployment diagnostics on Bit Insertion.}
  \textbf{(a)} The intervention ratio decreases over deployment, indicating
  reduced human-supervision burden.
  \textbf{(b)} TIDE achieves higher normalized task progress than variants
  without provenance conditioning or using only human-executed online data,
  highlighting the value of source conditioning and learning from
  policy-executed experience.}
  \label{fig:deployment-diagnostics}
\end{figure}

\paragraph{Q3: How does observed supervision burden evolve during deployment}
For each HIL training episode, we compute the intervention ratio as the
fraction of executed control steps under human takeover. In
\cref{fig:deployment-diagnostics}(a), gray markers show the per-episode
ratio for Bit Insertion, and the red curve shows the frame-weighted ratio
over the most recent five committed episodes. The observed windowed ratio
decreases after the early deployment phase, indicating lower supervision
burden over the session. This trace is not an autonomous-performance
measure and does not by itself isolate learning from temporal factors.

\paragraph{Q4: How important is provenance-conditioned learning}
\Cref{fig:deployment-diagnostics}(b) isolates the role of provenance
conditioning on Bit Insertion. All
variants start from the same warm-start checkpoint, use the same online
interaction budget, and are evaluated with 20 no-takeover rollouts from
the same initial states. TIDE obtains 80.0\% milestone-normalized task
progress, compared with 65.0\% without provenance conditioning and
67.5\% when only human-executed online chunks are used. This single-task
ablation supports both explicit source conditioning and the use of
policy-executed chunks for Bit Insertion.

\section{Conclusion}
\label{sec:conclusion}

We presented TIDE, a human-in-the-loop online post-training system that
turns real-robot deployment into a learning process for pretrained
manipulation policies. TIDE places robot execution, human takeover,
online data formation, joint optimization of all trainable policy
modules, and checkpoint handoff in one deployment session. Its
provenance-labeled chunks distinguish human-intervention from
policy-executed behavior by source, enabling supervised online training
without learned critics, trajectory-level reward labels, or action
scorers. Across four real-world tasks, TIDE achieves higher
milestone-normalized task progress and binary completion than the
warm-start and additional-offline-data baselines. Its final performance
is comparable to the Fixed Policy baseline, while updated checkpoints
become available during deployment. The observed intervention ratio also
declines over the Bit Insertion session, indicating lower supervision
burden rather than serving as a separate measure of autonomy.

\section{Limitations}
\label{sec:limitations}

TIDE requires a human supervisor, and retaining only manually confirmed
successful episodes may introduce selection bias. Our evaluation covers one
bimanual platform, four tasks, and one world-action backbone, and does
not test long-term retention or sensitivity to the online-training
schedule. Generalization across platforms and architectures therefore
remains unverified. We also do not measure variation across repeated online
post-training sessions.

\bibliographystyle{IEEEtran}
\bibliography{main}

\end{document}